# Radiuma: A Unified Zero-Code Executable Graphical Workflow Generator for Reproducible and Shareable Medical Image Analysis and Machine Learning

Mohammad Salmanpour[1,2,3*], Mehrdad Oveisi[3,4], Isaac Shiri[5,6], and Arman Rahmim[1,2,7]

[1]Department of Basic and Translational Research, BC Cancer Research Institute, Vancouver, BC, Canada
[2]Department of Radiology, University of British Columbia, Vancouver, BC, Canada
[3]Technological Virtual Collaboration (TECVICO Corp.), Vancouver, BC, Canada
[4]Department of Computer Science, University of British Columbia, Vancouver, BC, Canada
[5]Department of Cardiology, Inselspital, Bern University Hospital, University of Bern, Bern, Switzerland
[6]Department of Digital Medicine, University of Bern, Bern, Switzerland
[7]Departments of Physics & Biomedical Engineering, University of British Columbia, Vancouver, BC, Canada

**(*) Corresponding Author:**

Mohammad Salmanpour, PhD

Department of Basic and Translational Research,
BC Cancer Research Institute,
Vancouver BC, Canada
**Email:** msalman@bccrc.ca

**Abstract**

Medical image computing software development is crucial for identifying novel imaging biomarkers in clinical research, with the potential to significantly enhance patient outcomes by enabling more accurate diagnosis, prognosis, and treatment strategies. A lack of standardized and user-friendly software environments has impeded their widespread adoption. This work presents Radiuma to address this challenge. The modular software, freely available to the community, design provides a flexible platform for reliable, user-friendly image reading across various modalities and formats; image viewer; image registration; image fusion; image processing; image segmentation; radiomics feature extraction; and machine learning (classification, regression, and clustering) model development. Radiuma offers a modular approach to medical image analysis, allowing users with varying levels of expertise to execute each module independently or combine them with other components. The output of one module can be graphically fed into another, enabling users to create custom, executable, and reproducible workflows with multiple steps. Further, the results of each module execution can be viewed directly in the visualization window, providing immediate feedback on the accuracy and effectiveness of each step. This modular approach provides flexibility and customization options for researchers and clinicians, enabling reliable, reproducible radiomics research in clinical practice settings. Furthermore, Radiuma offers a convenient way to save and share customized workflows with other users, promoting reproducibility, reusability, and transparency in medical image and machine learning analysis. This feature is particularly useful for collaborative research projects and for ensuring consistency across different studies. This key feature and user-friendliness make it accessible for researchers at various skill levels, including radiologists, physicists, and data scientists.

## 1. Introduction

The development of medical image computing software plays a critical role in identifying novel imaging biomarkers within clinical research[1,2], with the potential to enhance patient outcomes by providing more accurate diagnostic procedures and optimizing therapeutic interventions[3,4]. Imaging biomarker development is a multidisciplinary, multistage process that requires integrating diverse computational modules, each designed to support specific tasks essential to the identification, validation, and application of appropriate biomarkers[5]. Various imaging modalities, such as Ultrasound, X-ray, Magnetic Resonance Imaging (MRI), Computed Tomography (CT), Positron Emission Tomography (PET), Single Photon Emission Computed Tomography (SPECT), and multi-modality systems (e.g., PET/CT, PET/MRI, SPECT/CT, or other fused modalities) are routinely used in clinical practice. Each imaging modality, with its unique physics, provides specific information, ranging from anatomical structure to functional and metabolic activity, and can be used independently or integrated to uncover novel imaging biomarkers for personalized medicine[6,7].

The development of imaging biomarkers from diverse modalities using medical image computing software involves multiple stages and tasks[5]. It is crucial for medical imaging software to support multiple imaging modalities, each with distinct image formats from different vendors. Additionally, the software must offer advanced visualization capabilities, including an interactive user interface and intuitive visualization paradigms, to enable users to effectively view and interact with medical images[1]. To quantify regions/volume of interest (ROI/VOI), the software should facilitate efficient image visualization and provide robust tools for segmenting these regions[8]. Furthermore, tasks such as image alignment, including registration and fusion of images from the same modality acquired at different time points or sequences, or from entirely different imaging modalities, may be necessary at various stages of analysis[9]. The software should also be capable of extracting quantitative metrics, including intensity and volumetric information of ROI or VOI, advanced radiomics features, and deep learning imaging features across multiple modalities and clinical settings[6]. Moreover, integrating advanced analytical techniques, such as machine learning, is essential to support a wide range of tasks, including biomarker discovery, predictive modelling, and clinical decision-making, thereby enhancing the potential of personalized medicine.

Generalized medical image computing software for developing imaging biomarkers should provide a flexible infrastructure that can adapt to various imaging modalities, support multiple image formats, and perform a wide range of computational tasks. Various open-source software tools have been developed by the medical imaging and machine learning communities to support image processing and machine learning studies. For medical image processing, open-source software such as ITK-SNAP[10], MANGO[11], MITK[12], 3D-Slicer[13], and CaPTk[14] have been developed to perform tasks such as image visualization, segmentation, registration, and quantification. For machine learning, open-source libraries such as scikit-learn[15] (which requires coding expertise) and user-friendly platforms such as WEKA[16] (which does not require coding) have been developed to facilitate general-purpose machine-learning modeling and analysis. In the field of radiomics, a variety of tools have been introduced, including six libraries, two command-line tools, two hybrid library/command-line tools, and 13 graphical user interface-based software (web and stand-alone)[17]. These tools are primarily designed for radiomics feature extraction, with only 5 supporting advanced analyses through modelling[17]. One of the most recent advancements is QuantImage v2, a web-based application that supports radiomics feature extraction and machine learning-based classification[17].

Despite the availability of these tools, none are designed to support both image processing and machine learning tasks comprehensively. Researchers often need to use multiple software tools to complete different stages of their workflow, which reduces efficiency, reproducibility, and reliability. This limitation is particularly problematic in imaging biomarker development and radiomics studies, where small changes in software versions or libraries can lead to significant variations in results. High inter- and intra-study variability further complicates the reproducibility of findings. Although guidelines have been developed to standardize specific aspects of image processing and machine learning workflows, such as radiomics feature standardization, imaging biomarker validation, and machine learning processes, these guidelines often focus on narrow domains[18-21]. Even when followed, they do not guarantee fully reproducible results due to implementation inconsistencies across studies.

Reproducibility challenges have been successfully addressed in other fields, such as bioinformatics, through tools like Galaxy[22]. Galaxy introduces the concept of workflows, allowing researchers to create, save, and share reproducible workflows with others (from data loading to final analysis)[22]. In medical imaging, radiomics, and machine learning studies, efforts to improve reproducibility have included sharing code or settings for specific tasks (e.g., radiomics feature extraction or machine learning modelling). However, these efforts lack the ability to share the entire workflow, which is critical for ensuring consistent and reproducible results across studies. The lack of integrated tools that support the entire image-processing and machine-learning pipeline, combined with the absence of standardized workflows, limits the reproducibility and reliability of image-biomarker development and radiomics studies. Addressing these challenges requires the development of comprehensive platforms that enable end-to-end workflow management, similar to the workflow concept in bioinformatics[22], to enhance reproducibility and facilitate collaboration in medical imaging research.

The lack of standardized, user-friendly, specialized software environments for generating medical imaging workflows has significantly hindered the widespread adoption of imaging biomarkers and radiomics studies. To address this challenge, we present Radiuma, an open-access, all-in-one, zero-code, executable, shareable, and reproducible medical imaging workflow generator for image processing and machine learning. This user-friendly modular software provides a flexible platform for reliable and easy-to-use image reading from different modalities and formats, image conversion, image registration, image fusion, image processing, image visualization, image segmentation, radiomics feature extraction, and machine learning model development for classification, regression, and clustering. Moreover, all these steps for a specific project can be executed independently or within a workflow-based approach. These workflows can be saved, reloaded, reconfigured, or transferred to other platforms, ensuring reproducibility and flexibility. Additionally, workflows can be shared with other researchers, promoting collaboration and reducing the time required to replicate or modify processes. By overcoming the limitations of previous software, Radiuma aims to promote the broader adoption of imaging biomarker development in clinical practice and research. Its intuitive design and zero-code approach make it accessible to researchers and clinicians with varying levels of expertise, enabling advanced imaging studies without requiring any programming knowledge.

## 2. Materials and Methods

**Figure 1** presents the schematic overview of the Radiuma software, including its modules and corresponding functionalities.

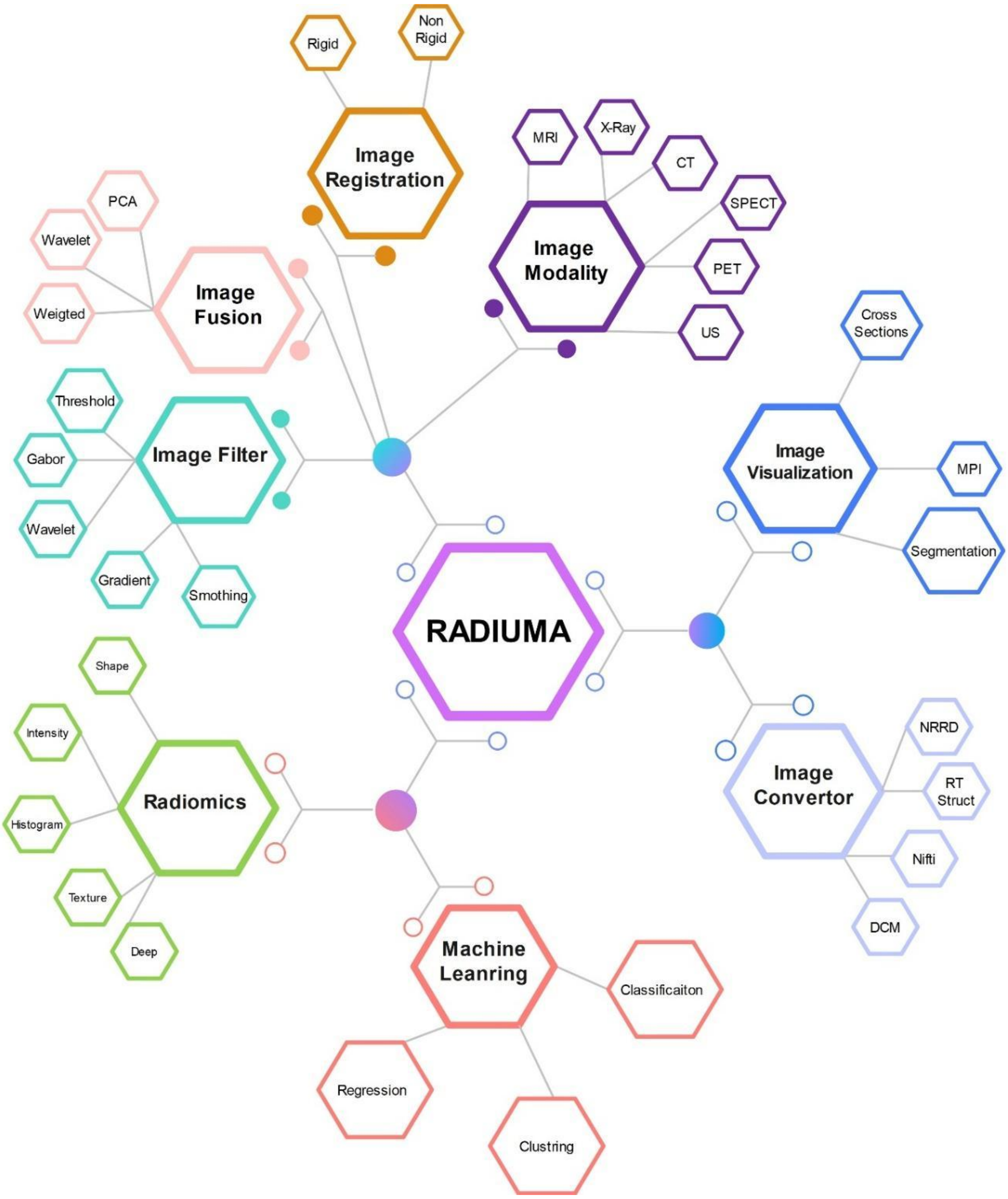

**Figure 1:** Schematic overview of the Radiuma software, including its modules and corresponding functionalities.

**Image Modalities and Format Support**

Radiuma supports all major imaging modalities, such as Ultrasound, X-ray, CT, MRI, SPECT, and PET, and is compatible with a wide range of formats, including DICOM, NIfTI, NRRD, and RT structures. Also, it provides an image conversion tool that facilitates transformation between different formats across different imaging modalities. The conversion functionality supports both single image series and batch processing, ensuring efficient handling of large datasets.

**Image Viewer**

Radiuma provides image viewer tools that enable users to view images in multiple views, including 3D (sagittal, axial, and coronal planes), maximum-intensity projection (MIP), and multimodal image fusion. These capabilities allow users to explore medical images from various perspectives, facilitating a more comprehensive understanding. Furthermore, Radiuma's 3D visualization tools allow users to rotate and zoom in and out, providing enhanced control and flexibility during the visualization process. Additionally, the software allows users to overlay segmented regions onto the original images, offering a visual representation of the segmented areas within the context of the original anatomical structures.

**Image Segmentation**

Radiuma offers various image segmentation tools for delineating regions or volumes of interest (ROIs/VOIs), including manual segmentation, thresholding, and polygonal algorithms. With these tools, Radiuma enables users to delineate VOIs or ROIs. The diverse selection of tools provides users with the flexibility to choose the most suitable method for their specific requirements, ensuring optimal segmentation performance across various applications. Segmentations can be saved in various formats and utilized for further analysis, such as image quantification. Additionally, Radiuma supports RT-Structure, enabling the import and export of RT-Structure data. It also enables conversion between RT-Structure and other formats, such as DICOM or NIfTI, enhancing interoperability and flexibility in data handling.

**Image Registration and Fusion**

Radiuma provides users with a comprehensive suite of image registration options, offering varying degrees of freedom to accommodate diverse medical image needs. These registration capabilities enable alignment of multimodal images acquired at different time points, from different sequences of the same imaging modality (e.g., T1 and T2 MRI), from various imaging modalities (e.g., CT, MRI, PET, or SPECT), or even from different patients. Supporting rigid, affine, and deformable registration techniques ensures precise spatial alignment. This functionality is particularly valuable for longitudinal studies and multi-modal imaging workflows. In addition to image registration, Radiuma offers a variety of image fusion algorithms that facilitate the integration of data from multiple imaging modalities. These algorithms include weighted fusion, wavelet fusion, and principal component analysis fusion. By providing these advanced fusion techniques, Radiuma enables users to integrate data from diverse modalities seamlessly.

**Image Pre-Processing**

Radiuma offers a comprehensive array of image convolutional filters, standardized according to the Image Biomarker Standardization Initiative (IBSI 2.0)[19], for image pre-processing. These filters include the mean filter, Laplacian-of-Gaussian filter, Laws' kernels, Gabor kernels, separable and non-separable wavelets, and Riesz transformations. These filters enhance image quality, reduce noise, and enable the extraction of additional radiomics features from filtered images, thereby optimizing image preparation for subsequent analytical processes.

**Radiomics Feature Extraction**

Radiuma supports both handcrafted and deep radiomics feature extraction. The handcrafted feature extraction module in Radiuma is based on our previous open-source package, PySERA[23], which adheres to the IBSI 1.0 guidelines[18], and in combination with the convolutional filters, adheres to IBSI 2.0[19]. This enables the extraction of a wide range of standardized 2D and 3D handcrafted radiomics features from ROIs/VOIs, including shape, intensity, histogram, and texture features (first-order, second-order, and higher-order). Additionally, Radiuma integrates state-of-the-art deep learning algorithms, such as pretrained autoencoders and convolutional neural networks (CNNs), to extract deep learning-based radiomics features, further enhancing its capabilities for comprehensive radiomics analysis[23].

**Machine Learning**

Radiuma offers a comprehensive set of supervised, semi-supervised, and unsupervised machine learning algorithms designed for diagnostic, predictive, and prognostic modelling. Its pipelines encompass a wide range of functionalities, including feature pre-processing, dimensionality reduction, feature selection, and advanced modelling techniques such as regression, classification, and clustering. By providing a diverse array of algorithms for each stage of various machine learning pipelines, Radiuma enables users to analyze their data with exceptional flexibility. The machine learning pipelines in Radiuma can be integrated into workflows for radiomics analysis or applied as standalone tools to datasets from diverse sources. This dual functionality enables users to leverage Radiuma's advanced algorithms for both radiomics-based tasks and broader data analysis applications, making it a versatile solution for a wide range of research and clinical needs. Whether working with imaging-derived radiomics features or any external tabular datasets, Radiuma provides the tools to extract meaningful insights and drive data-driven decision-making.

**Workflow Generation**

Radiuma offers a streamlined workflow that seamlessly connects various modules to handle different tasks efficiently (**Figure 2**). For instance, images and their corresponding segmentations can be automatically loaded in various formats and converted to the desired format. These images can then be linked to image-processing modules for preprocessing with different filters. Following this, the radiomics feature extraction module can be configured to extract specific handcrafted or deep-learning features. The extracted features, along with clinical data and target variables, can be organized into tabular datasets. After preprocessing the tabular data, it can be passed to the machine learning module, where users can set parameters and automatically build, train, and evaluate various models to obtain results. This entire workflow can be saved and reloaded at any time to resume the process or reconfigured as needed. Additionally, the workflow can be exported and shared with others to ensure reproducibility of results, fostering collaboration and transparency in research and clinical applications. This integrated approach simplifies complex analyses, making Radiuma a powerful and user-friendly tool for advanced data processing and modeling.

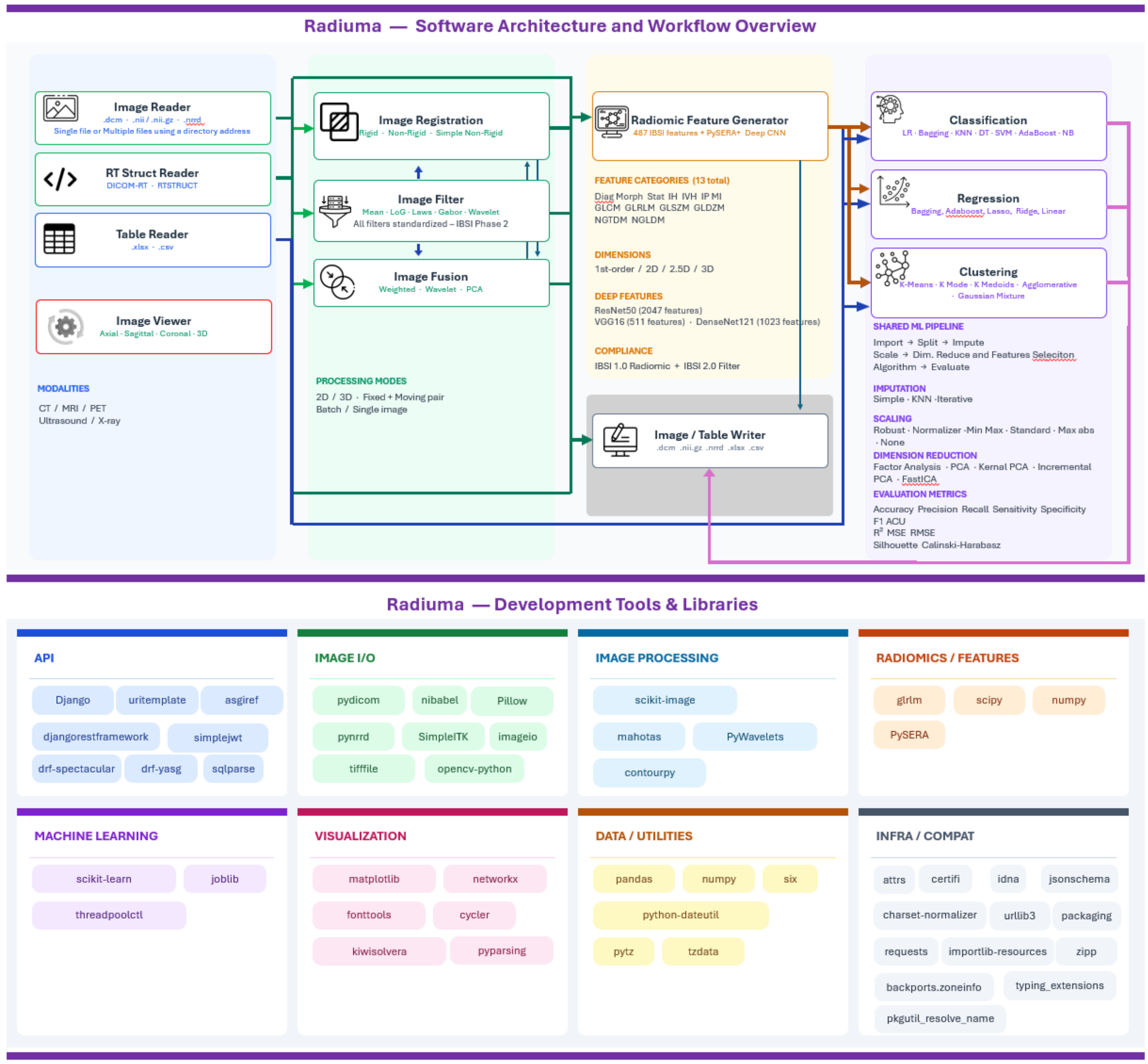

**Figure 2:** Overview of the Radiuma architecture. The upper panel shows the different modules of Radiuma and how they interact with each other within an end-to-end workflow, from image input through feature extraction and machine learning to result export. The lower panel shows the libraries and development tools used to build Radiuma.

**Software Development**

Various open-source Python libraries, including ITK, OpenCV, and scikit-learn, were used for software development. The software is compatible with multiple operating systems, including Windows, macOS, and Linux.

**Analysis**

We included datasets spanning multiple medical imaging modalities, including ultrasound, MRI, CT, and PET/CT, to evaluate different modules and create the documentation. Furthermore, we evaluated clinically relevant prediction targets, including isocitrate dehydrogenase (IDH)-wildtype mutation status prediction across different MRI sequences for radiogenomic analysis in glioblastoma (GBM)[24] in a full pipeline (image processing to machine learning modeling). We tested and documented the performance of various Radiuma modules across the clinical tasks outlined above.

## 3. Results

Radiuma incorporates a visual node-based workflow system that enables users to construct complex medical image analysis and machine learning pipelines through an intuitive, programming-free interface (**Figure 3**). Tools are represented as modular nodes that can be added, configured, and interconnected to form complete processing chains, with workflows executed either at the individual node level or globally. The platform supports simultaneous multi-workflow management through a tab-based interface, allowing users to create, rename, and load workflows independently. Tool compatibility is enforced through predefined input–output rules, ensuring valid pipeline construction across imaging processing, radiomics, preprocessing, and machine-learning modules. Advanced workflow controls, such as automated layout alignment, tool search, node configuration copy–paste, and full workflow save/load functionality, enhance reproducibility and efficiency. System resource monitoring alerts users to low memory or cache availability, while integrated keyboard shortcuts streamline interaction, enabling a robust, user-friendly environment for high-throughput medical image analysis. More detailed explanations and examples are provided in the supplementary dataset, along with documentation of Radiuma.

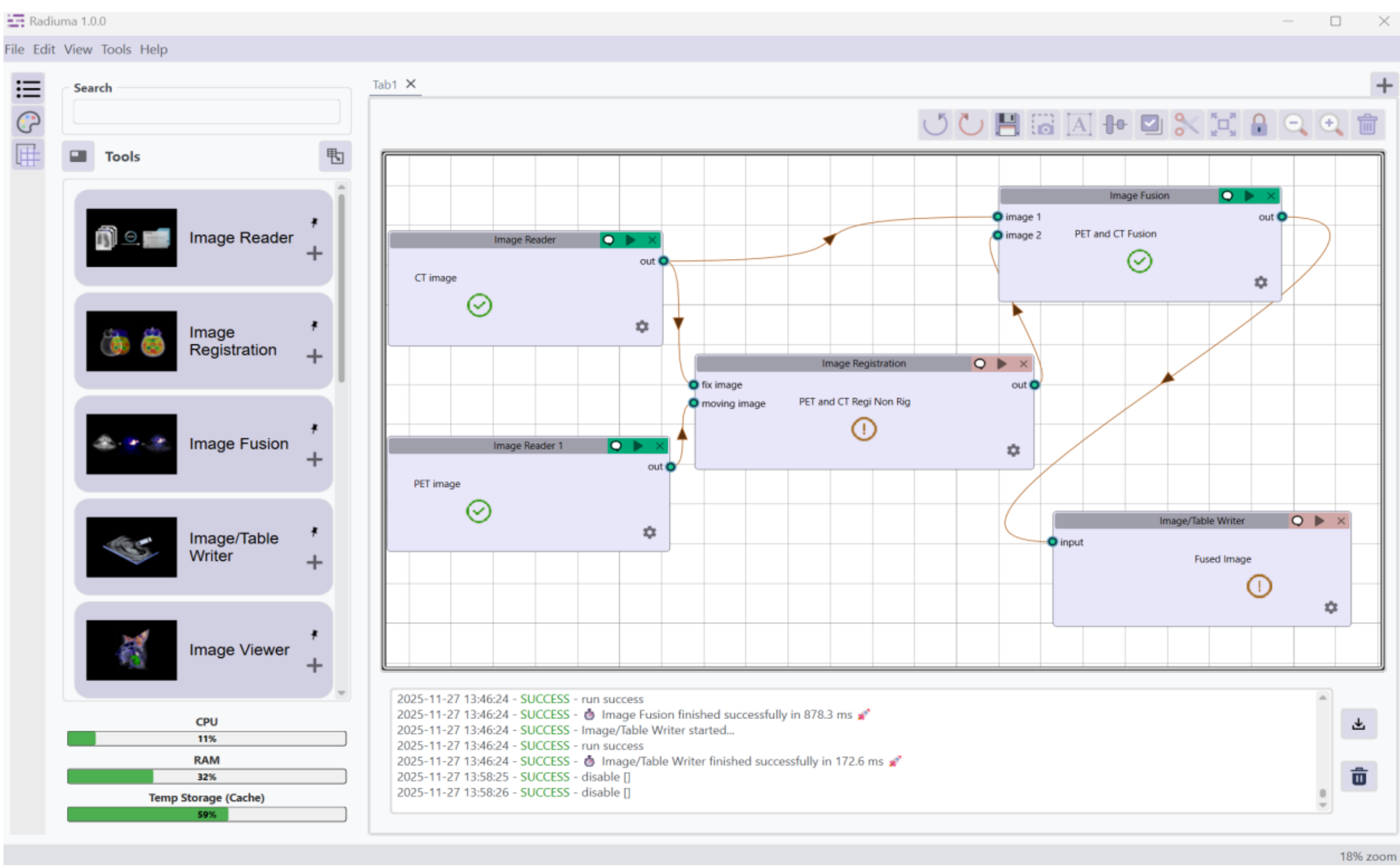

**Figure 3:** Radiuma interface demonstrating the node-based workflow editor and tab-based multi-workflow management, with multiple analysis modules connected into a complete medical image processing and radiomics pipeline.

### Image Viewer and Segmentation Tools

The Radiuma Image Viewer provides an advanced, multi-dimensional visualization environment designed for comprehensive inspection, manipulation, and analysis of medical imaging data (**Figure 4**). It supports 2D and fully rendered 3D views, enabling precise anatomical localization and interactive exploration of complex volumetric datasets. Key functionalities include interactive coordinate and intensity readouts, contrast and colormap control, filtering, rotation, cropping, distance measurement, overlay visualization, and synchronized crosshair navigation across all views. Advanced segmentation capabilities are provided through manual and threshold-based drawing tools, label creation, volume-mask toggling, and editable ROIs, supported by brush and eraser operations. Additional features such as multi-view breeding, screenshot capture, and detailed metadata access further enhance usability. More detailed explanations and examples are provided in the supplementary dataset, along with documentation of Radiuma.

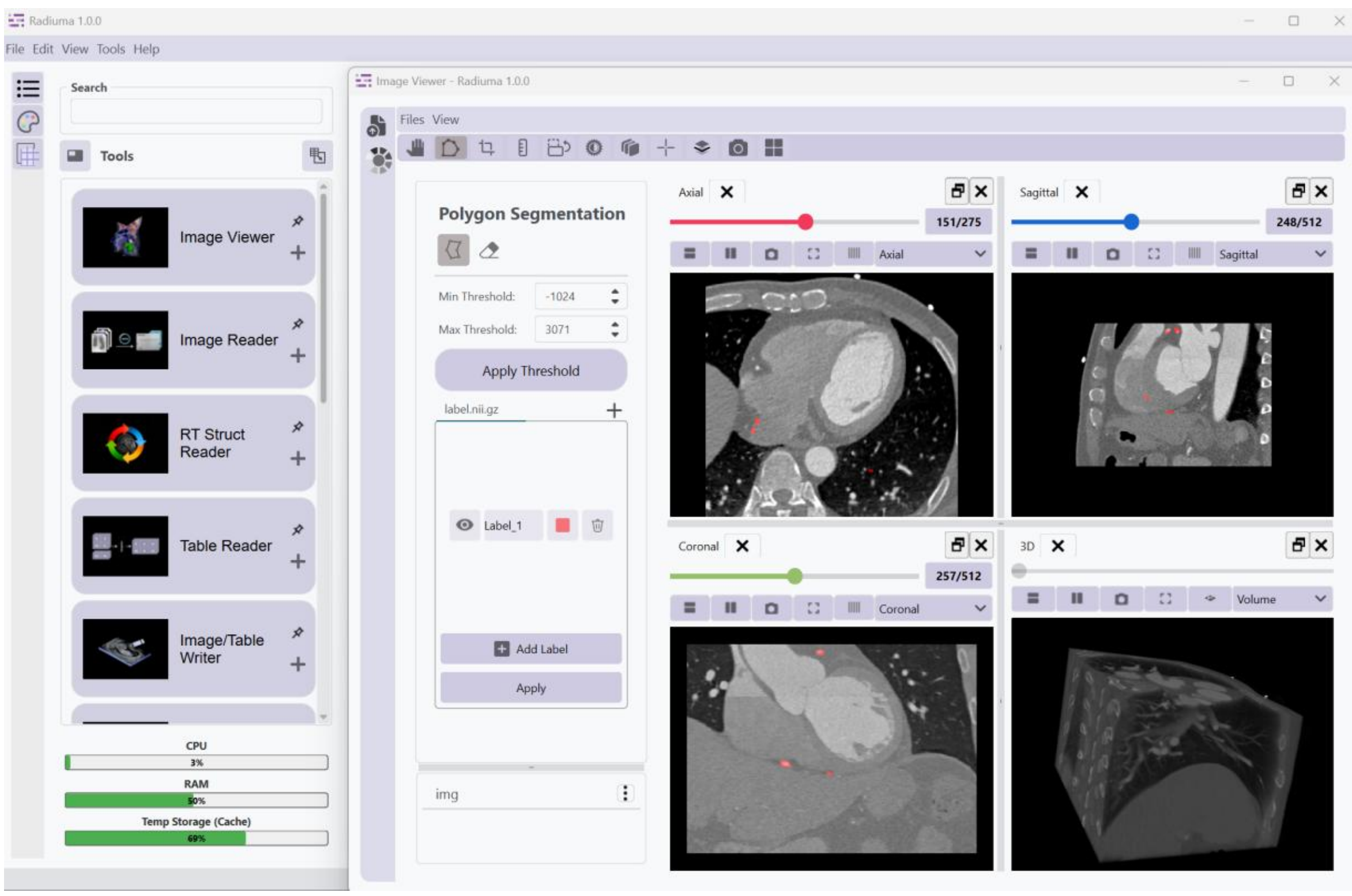

**Figure 4:** Radiuma Image Viewer demonstrating multi-planar and 3D views of medical imaging data with segmentation overlays. The viewer supports axial, sagittal, coronal, and volumetric renderings, along with tools for contrast adjustment, filtering, measurement, cropping, and segmentation editing.

### Image, RT Struct, Table reader and Writer Tools

Radiuma provides a streamlined suite of data Input/Output modules for importing, organizing, and exporting diverse imaging and tabular datasets within medical analysis and machine learning workflows (**Figure 5**). The Image Reader supports DICOM directories, individual DICOM files, NIFTI volumes, and NRRD files, and automatically detects patient- and modality-specific structures to ensure proper dataset organization. The RT Struct Reader enables seamless integration of DICOM-RT structure sets by pairing contour files with their corresponding images, supporting radiomics and dosimetric analysis. For structured data, the Table Reader imports CSV, Excel, and other formats and allows row- and column-wise merging to create unified datasets combining clinical and imaging-derived variables. The Image/Table Writer provides flexible export options for processed images and analysis outputs in standardized formats. More details are available in the supplementary material and Radiuma documentation.

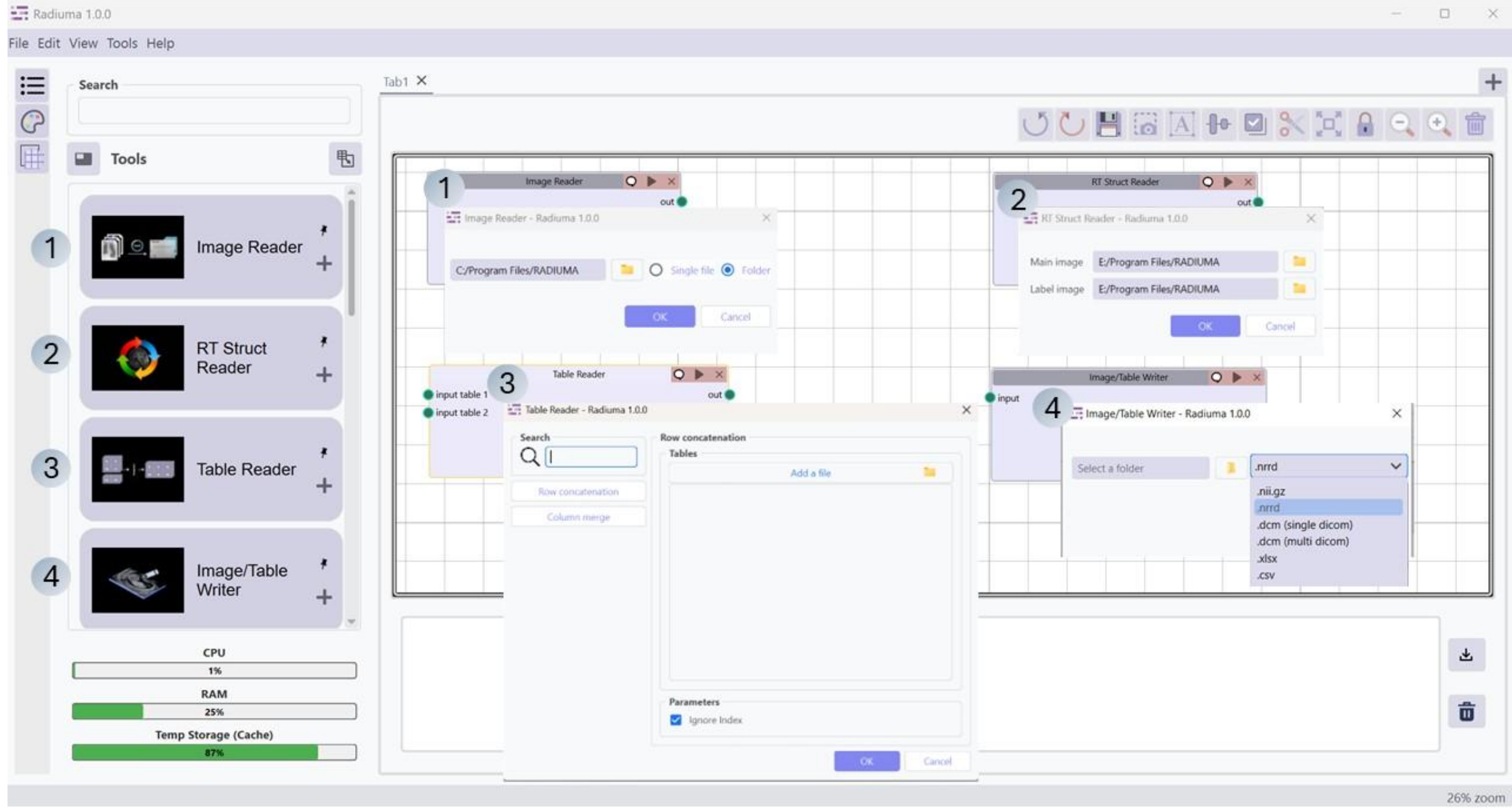

**Figure 5:** Overview of Radiuma modules: (1) Image Reader, supporting single-file loading as well as batch reading of all files within a directory; (2) RT-Struct Reader, enabling the import and conversion of images and corresponding segmentation labels; (3) Table Reader, providing options for column merging and row concatenation; and (4) Image/Table Writer, supporting export to NIfTI, NRRD, and DICOM (single or multi-file), as well as Excel and CSV formats.

### Image Registration and Fusion Tools

The registration tool supports both rigid and non-rigid methods, enabling accurate alignment across modalities and time points and offering configurable optimization, interpolation, and deformation settings. The fusion module integrates complementary anatomical and functional information using different techniques and requires intensity normalization to ensure reliable results. Together, these modules provide robust preprocessing capabilities that enhance multimodal image comparison, improve anatomical consistency, and enable enriched fused representations for downstream analysis. **Figure 6** illustrates the registration and fusion of PET and CT images using the workflow described in **Figure 3**, along with their visualization in the image viewer modules.

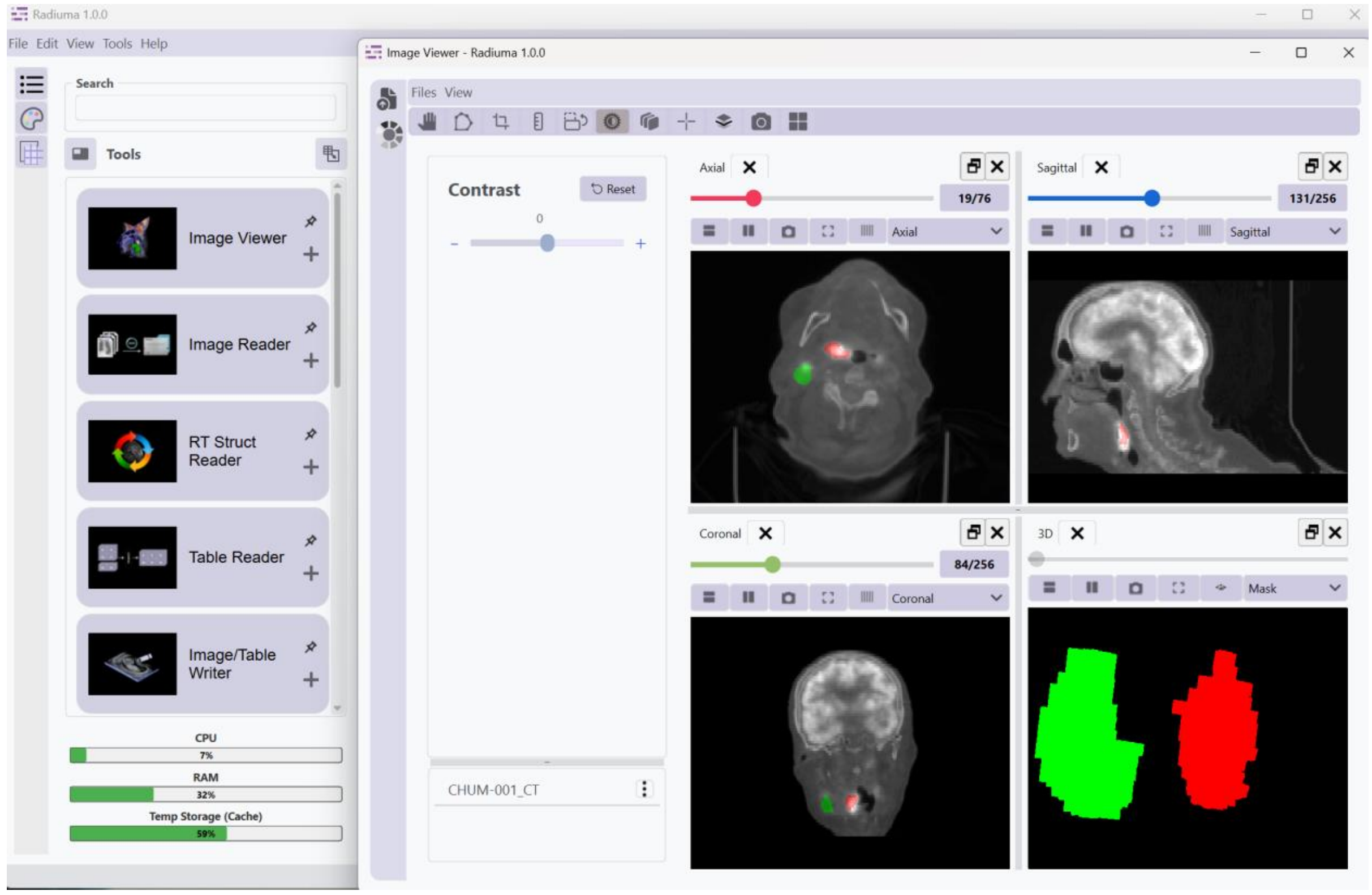

**Figure 6:** Example of PET and CT images loaded and registered using non-rigid registration (default parameter of Radiuma), followed by weighted fusion (z-score normalization with 50% weighting for each modality and nearest-neighbor interpolation). The fused image is displayed in the image viewer, with tumor regions shown in red and lymph nodes in green.

### Radiomics Feature Generator Tool

The Radiomic Feature Generator in Radiuma provides a fully standardized, comprehensive framework for quantitative imaging analysis, enabling the extraction of handcrafted and deep learning–based features from multimodal medical images. Powered by the IBSI-1–compliant PySERA[18,23] backend, the module supports the computation of 557 features, including 487 IBSI-compliant radiomic descriptors, 60 diagnostic metrics, and 10 moment-invariant features across 1st-order, 2D, 2.5D, and 3D spatial dimensions[23]. The tool offers flexible configuration options, such as modality-specific preprocessing, ROI selection strategies, multi-lesion feature aggregation, and advanced handling of missing feature values. In addition to traditional radiomics, the module extracts high-dimensional deep features from pretrained CNN architectures (ResNet50, VGG16, and DenseNet121), providing feature vectors ranging from 511 to 2047 elements[23]. Integrated with standardized filtering, discretization, and resampling settings, the radiomics pipeline produces harmonized, reproducible feature tables directly compatible with statistical modeling and machine learning workflows[23]. Overall, the Radiomics module provides a robust, reproducible quantitative imaging framework for disease characterization, biomarker discovery, and predictive modeling. Complementing this, the platform integrates an IBSI-2–compliant[19] Image Filter module that offers a broad suite of standardized preprocessing filters, including Mean, LoG, Laws, Gabor, and Wavelet, to enhance image quality and texture representation before feature extraction.

**Machine Learning Tools**

For labeled tasks, the Classification and Regression modules support structured data import with ID matching, advanced imputation, feature scaling, dimension reduction, hyperparameter tuning, and side-by-side evaluation of multiple algorithms (e.g., logistic regression, SVM, tree-based ensembles, linear and regularized regressors) using standard performance metrics. For unlabeled data, the Clustering module implements a similar preprocessing pipeline and offers several algorithms (K-means, K-medoids, K-mode, Gaussian mixtures, and Agglomerative) with systematic tuning and internal validation indices to identify stable patient subgroups or imaging-derived phenotypes. Together, these modules enable end-to-end model development, from raw feature tables to optimized, quantitatively assessed predictive or exploratory models, without requiring users to write code.

**Workflow Generation and Reproducibility**

Complete analytical workflows in Radiuma, covering image loading, preprocessing, segmentation, registration, radiomics extraction, and machine learning, were constructed using a graphical node-based interface, saved, reloaded, and executed across different operating systems (Windows, macOS, Linux). Workflows are built by adding and configuring compatible modules (e.g., Image Reader → Filter/Fusion/Registration, RT Struct Reader → Radiomic Feature Generator/Image Writer, radiomics → classification/regression/clustering) and connecting them via directed links, with execution controlled through node-level run/stop actions. Reloaded workflows produced identical feature matrices and model outputs across platforms and users, demonstrating full adherence to URRA (usability, reusability, reproducibility, and accessibility) principles and highlighting the advantages of Radiuma's executable Graphical User Interface (GUI)-based workflows over manually scripted pipelines. We have provided multiple workflows in the documentation.

**Analysis**

A full pipeline integrating image processing, registration, fusion, radiomics feature extraction, and machine learning was applied across multiple medical imaging datasets as documented in supplementary data and documentation. IDH-wt prediction in GBM using MRI reached for the T1CE modality, the combination of PCA for dimensionality reduction and logistic regression as the classifier yielded moderate classification performance. On the validation set, the model achieved an F1 score of 0.70 and an AUC of 0.76, suggesting reasonable discriminative ability during model selection (as the best model). However, a slight decline in performance was observed on the held-out test set, where the F1 score dropped to 0.62 while the AUC remained relatively stable at 0.77, indicating that the model retained acceptable generalizability despite some reduction in precision-recall balance.

## 4. Discussion

Imaging biomarker development is a multi-stage process involving experts across medical imaging, radiology, medical physics, and computer science[3]. It relies on diverse tools and software[3,4,6] to support tasks such as visualization, segmentation, quantification, registration, fusion, feature extraction, and diagnostic or prognostic modeling. Although many tools exist, a unified framework integrating multimodal imaging, radiomics, and machine learning for both clinical and research applications remains lacking. Such integration is essential for robust biomarker validation, reproducibility, and personalized medicine. To address this need, we have developed an open-access software suite with modules for comprehensive image processing and machine learning tasks. It enables the creation, saving, and sharing of workflows to promote reproducible imaging biomarker research. Radiuma is a zero-code, executable platform designed for standardized imaging biomarker discovery.

While the dissemination of image processing and machine learning models and code has expanded significantly over the past decade, many shared resources continue to suffer from poor usability and limited reproducibility[25]. Reliable computational research in healthcare depends on tools and software that uphold the principles of URRA. These attributes not only strengthen scientific validity but also facilitate the integration of computational innovations into clinical workflows. Accordingly, URRA principles now occupy a prominent role in scientific methodology, particularly in medical imaging research[26-34].

Growing evidence highlights the critical role of reproducible environments and shareable workflows in ensuring scientific reliability[26-34]. Strikingly, a survey of 1,500 researchers revealed that more than 70% failed to reproduce others' experiments, and over 50% could not replicate their own work[35]. Reproducibility assessments reveal substantial variability across the literature. In one analysis of 255 papers published between 1984 and 2017, only 64–85% of results could be replicated, depending on whether the original authors assisted[36]. Similar large-scale studies report that nearly half of the 500 primary investigations were not reproducible[36-38]; one reproduced just 70% of 232 papers[39], and another confirmed reproducibility in only 9 of 25 articles[38]. Recent studies have demonstrated substantial variability when the same workflow is applied to the same dataset across different platforms[40]. Other recent work has reported inconsistent outcomes, even when identical workflows and datasets were used under the direct supervision of the original author[41]. Additional surveys[42-45] reinforce these concerns, highlighting a persistent reproducibility crisis that continues to hinder the translation of research into clinical practice.

Tools such as Docker, Git, R-Studio, and Airflow support URRA principles for general-purpose computational workflows, and platforms like Nextflow[46], Bioconductor, and Galaxy[22] extend similar capabilities within bioinformatics; however, none of these solutions address the specific URRA challenges of medical imaging[26-34]. At the same time, existing open-source tools for imaging and machine learning remain fragmented: software such as ITK-SNAP[10], MANGO[11], MITK[12], and 3D-Slicer[13], provide strong visualization, segmentation, and registration functionalities but lack advanced machine learning modules, while general-purpose platforms like WEKA[16] are not designed for medical imaging data. Radiomics tools such as QuantImage v2[17] focus mainly on feature extraction and offer limited modeling capabilities, with minimal support for multimodal workflows or broader image processing tasks. Collectively, these limitations show the need for a unified, URRA-compliant framework that integrates standardized imaging, radiomics, and machine learning to support comprehensive research and clinical pipelines.

URRA challenges are exacerbated by the fact that many medical imaging researchers and clinicians have strong domain expertise but limited programming experience, leaving them reliant on existing tools rather than writing custom code. Even when coding is possible, implementations rarely follow standardized practices and often vary across groups and projects, hindering reproducibility. Radiuma addresses these issues by offering an all-in-one, URRA-compliant platform for image analysis and machine learning workflows. Its modular graphical interface breaks complex processes into manageable components, enabling consistent, reusable, shareable, and accessible workflow design for users with varying technical skills and promoting more unified practices in imaging biomarker research.

Although we have developed a versatile, open-access, standardized platform for medical image analysis and machine learning, several limitations remain that we aim to address in future updates. For segmentation, we plan to integrate automated deep learning–based tools that allow users to import and apply pre-trained models across imaging modalities. In the machine learning module, we intend to add time-to-event modeling to support survival analysis, an essential aspect of clinical research. Additionally, while deep learning is currently restricted to feature extraction, we aim to extend its use to other modules, including segmentation, registration, and end-to-end model development, enabling users to apply deep learning throughout the entire workflow, from preprocessing to deployment.

## 5. Conclusion

Radiuma is an integrated medical imaging software platform, made publicly available to the community, designed to standardize image analysis and machine learning, and to streamline biomarker discovery for clinical research and clinical practice. Its user-friendly and modular architecture incorporates a broad range of image-processing and machine-learning algorithms, offering flexibility and customization for users with varying technical backgrounds. By enabling reliable, reproducible, and reusable analyses, Radiuma bridges the gap between complex computational workflows and practical clinical applications, supporting advances in medical research and advancing the development of personalized medicine.

**Code and Data Availability**

The developed software is freely available for download and use in various operating systems of Windows, macOS, and Linux, from the Radiuma website (https://radiuma.com/products/solutions/radiuma/). Comprehensive documentation with extensive examples is available on the website and on "Read the Docs": https://radiuma-documentation.readthedocs.io/en/ . The training videos are available to help users effectively use the software at https://www.youtube.com/@Radiuma-software .

**Acknowledgment**

The authors would like to acknowledge Sonya Falahati, Mohammad Ahmadi, Sirwan Barichin, Yasaman Salehi, and Amin Baghaei for their valuable support and *technical* assistance throughout this study.

**Funding**

We acknowledge funding from the Natural Sciences and Engineering Research Council of Canada (NSERC) Idea to Innovation (I2I) Grant GR034192 and Discovery Horizons Grant DH-2025-00119. This study was also supported by the Virtual Collaboration Group (VirCollab, *www.vircollab.com*) and Technological Virtual Collaboration (TECVICO CORP.), both based in Vancouver, Canada.

**Conflict of interest**

M. Salmanpour and M. Oveisi are affiliated with TECVICO Corp. and VirCollab, while the other co-authors have no relevant conflicts of interest to disclose.